\documentclass{article}
\usepackage{ijcai23}

\usepackage{times}
\usepackage{soul}
\usepackage{url}
\usepackage[hidelinks]{hyperref}
\usepackage[utf8]{inputenc}
\usepackage[small]{caption}
\usepackage{graphicx}
\usepackage{amsmath}
\usepackage{amsthm}
\usepackage{booktabs}
\usepackage{algorithm}
\usepackage{algorithmic}
\usepackage[switch]{lineno}

\usepackage{hyperref}
\usepackage{color}
\usepackage{amsthm,amsmath,amssymb}
\usepackage{subfigure}
\usepackage{mathrsfs}
\usepackage{booktabs}
\usepackage{multirow}
\usepackage{amssymb,mathrsfs,amsmath}
\title{Low-Rank Winograd Transformation for 3D Convolutional Neural Networks}

\author{
Ziran Qin$^1$
\and
Mingbao Lin$^2$\and
Weiyao Lin$^{1}$\thanks{Corresponding Author}%\And
\affiliations
$^1$Shanghai Jiao Tong University\\
$^2$Tencent Youtu Lab%\\
\emails
qinziran@sjtu.edu.cn,
linmb001@outlook.com,
wylin@sjtu.edu.cn
}
\begin{document}

\maketitle
\begin{abstract}
This paper focuses on Winograd transformation in 3D convolutional neural networks (CNNs) that are more over-parameterized compared with the 2D version. The over-increasing Winograd parameters not only exacerbate training complexity but also barricade the practical speedups due simply to the volume of element-wise products in the Winograd domain. We attempt to reduce trainable parameters by introducing a low-rank Winograd transformation, a novel training paradigm that decouples the original large tensor into two less storage-required trainable tensors, leading to a significant complexity reduction. Built upon our low-rank Winograd transformation, we take one step ahead by proposing a low-rank oriented sparse granularity that measures column-wise parameter importance. By simply involving the non-zero columns in the element-wise product, our sparse granularity is empowered with the ability to produce a very regular sparse pattern to acquire effectual Winograd speedups. To better understand the efficacy of our method, we perform extensive experiments on 3D CNNs. Results manifest that our low-rank Winograd transformation well outperforms the vanilla Winograd transformation. We also show that our proposed low-rank oriented sparse granularity permits practical Winograd acceleration compared with the vanilla counterpart. 
\end{abstract}

\section{Introduction}
%
%

%Compared to 2D version, 
3D convolutional neural networks (CNNs) have achieved substantial accuracy increases in many video processing tasks, as a result of their superior capacity of extracting spatio-temporal features. 
%
%Unfortunately, 
%The supreme performance is gained at the price of large amounts of computing resources for both training and inference, primarily because the 3D kernels are more computationally intensive. 
%
Unfortunately, they come with large amounts of computing resources, 
%for both training and inference, 
primarily because the 3D kernels are more computationally intensive. 
Many restrictions such as runtime, memory, and power budget prevent 3D CNNs from running on many real-world devices.

Fast convolution algorithms such as Winograd convolution~\cite{lavin2016fast} and fast Fourier transform (FFT)~\cite{mathieu2013fast} can greatly reduce the convolutional cost. %computational cost of the spatial convolution.
%
%The principle of Winograd convolution and FFT
Their principle is to replace spatial convolution operations with element-wise product and discard redundant multiplications in convolution.
Conventional FFT based convolution is fast for large filters, therefore most recent attention focuses on Winograd convolution principally for the small 3$\times$3 filters adopted by state-of-the-art CNNs.
%
%Another potential line to tackle the over-parameterized issue 
Another potential solution is network pruning that reduces network complexity by removing unnecessary units~\cite{frankle2018lottery,luo2017thinet}. 
It seems that Winograd convolution and network pruning can be well combined to further save computation costs. 
However, they are not naturally compatible since the sparsity property from network pruning is diminished after the kernel transformation of the Winograd algorithm.
To tackle this incompatibility, \cite{li2017enabling} performed pruning operations upon Winograd domain while~\cite{liu2018efficient} added the ReLU function after the Winograd transformation to increase the sparsity of element-wise product.
However, both studies do not 
%take into account a reality 
consider that Winograd kernel is location-sensitive as later demonstrated by~\cite{li2017enabling} where an importance factor matrix is utilized to gauge the significance of different kernel locations.

\begin{figure*}[t]  %子图加并列
\centering
\includegraphics[width=0.92\textwidth]{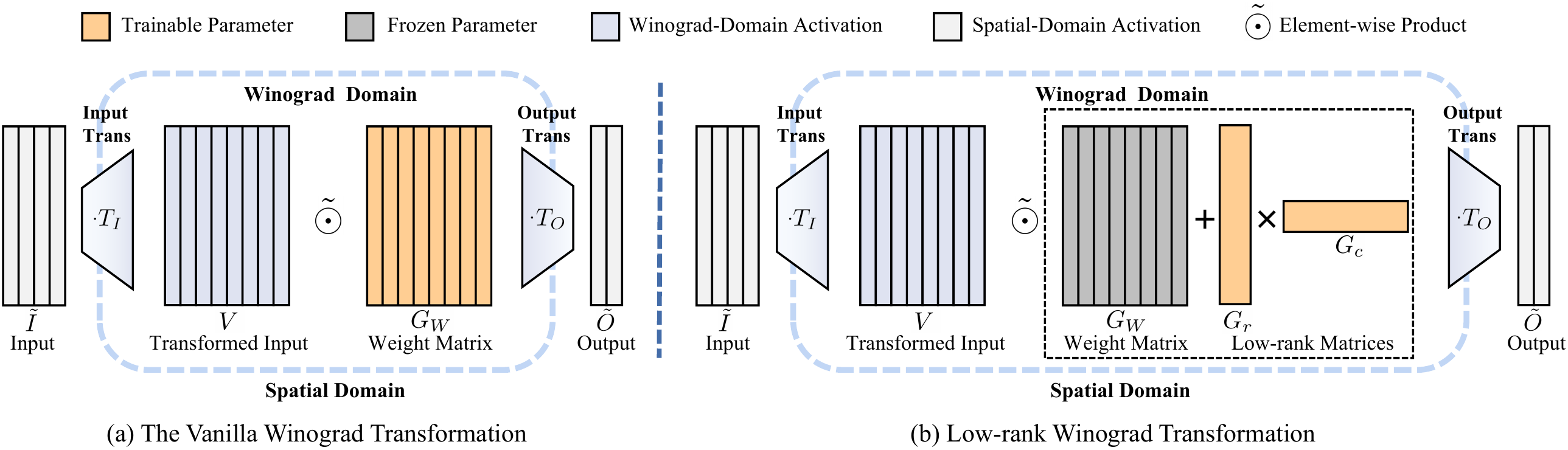} 
\caption{%The framework of training the 3D Winograd layer. (a) Training with full Winograd weights. (b) Training with proposed $XXX$.
Comparison between (a) the vanilla Winograd transformation and (b) our low-rank Winograd transformation. We decouple the whole Winograd weights into two smaller matrices, leading to a significant reduction in trainable parameters.
}
\label{fig:inp}
\end{figure*}

%
%In spite of the aforementioned progress, 
%
Current investigations mostly give attention to the Winograd transformation in 2D CNNs. A direct extension of these methods to 3D CNNs is inapplicable, as we analyze, for two issues.
First, 3D Winograd transformation causes considerable parameter increase. Taking F(2, 3)-based Winograd algorithm as an example, a typical 3D convolutional kernel with a shape of $3\times3\times3$ is often replaced by a $4\times4\times4$ Winograd kernel, leading to 2.37$\times$ more parameters while it is only 1.78$\times$ for 2D case. %
More parameters from Winograd transformation do not always benefit model capacity but cause model redundancy as analyzed in Sec.\,\ref{low-rank} and verified in Sec.\,\ref{performance_result}.
Also, the over-increasing trainable parameters pose a serious challenge to the capability of training machine.
Second, existing methods fail to accelerate Winograd transformation even though conducting pruning upon the Winograd domain. Similar to the weight pruning~\cite{lecun1989optimal,han2015learning,frankle2018lottery}, prior implementations derive irregular sparse weight matrix, which receives very limited speed gains since the irregular sparsity barely takes advantage of vector processing architectures such as single instruction multiple data (SIMD), and poorly utilizes memory buses~\cite{lin20221xn}.
Therefore, it remains unsolved to excavate Winograd transformation for acceleration, in particular to 3D CNNs primarily for the ever-increasing element-wise product in the Winograd domain.

In this paper, we put forward a novel Winograd transformation for 3D CNNs towards solving the above issues. 
Considering the over-increasing parameters in 3D CNNs, as shown in Fig.\,\ref{fig:inp}, compared with the vanilla version, we introduce a low-rank Winograd transformation method that represents the updating matrix (variation from a pre-trained Winograd weight tensor to the final fine-tuned one) with two smaller matrices. In this fashion, we concentrate on updating weights in the main directions of the whole Winograd space during the sparse training, leading to superior performance over the vanilla Winograd transformation. Besides, the two less storage-required matrices lead to a significant reduction in trainable Winograd parameters. 
With regard to Winograd transformation acceleration, we further present a low-rank oriented sparse granularity that quantifies the importance of each tensor column. The rationale behind this is to derive a more regular sparse pattern by simply involving the non-zero columns in the element-wise product of the Winograd domain. Thus, we introduce a scoring sequence to continuously accumulate the magnitude and gradient of column location in each training iteration as the importance assessment, and finally remove all weights in compliance with the low-scored columns. Practical speedups are observed from our low-rank oriented sparse granularity (see Table\,\ref{speedtest}).

\section{Related Work}
\subsection{Spatial-Domain Pruning}
Spatial-domain pruning is the practice of removing parameters from an existing network. It may entail removing individual parameters, \emph{a.k.a.} weight pruning, or parameters in groups such as filter pruning and block pruning. 
For weight pruning, individual weights are measured by a certain criterion such as weight magnitude~\cite{han2015learning,han2015deep,frankle2018lottery}, higher-order information~\cite{lecun1989optimal,hassibi1993optimal,dong2017learning,lee2018snip} and so on.
These methods are demonstrated to well preserve model performance. However, the resulting irregular sparse matrix requires specialized hardware/libraries to achieve practical speedups.
For filter pruning, the entire filters are removed by standards such as $\ell_1$/$\ell_2$-norm~\cite{li2016pruning,liu2017learning,he2018soft}, activation sparsity~\cite{hu2016network}, lasso regression-based channel selection~\cite{he2017channel}, and rank of feature maps~\cite{lin2020hrank}. In contrast to weight pruning, filter pruning advantages in acceleration but causes more performance drops.
Therefore, block pruning, where a block of weights is removed simultaneously, has received recent research focus~\cite{niu2021rt3d,lin20221xn} for its better performance than filter pruning as well as hardware-friendly deployment than weight pruning.
However, vanilla spatial pruning methods cannot be directly combined with the Winograd convolution because the Winograd transformation diminishes the sparsity resulting from pruning~\cite{yu2019spatial}.
\subsection{Winograd-Domain Pruning}
Though vanilla spatial pruning fails to cooperate with the Winograd convolution, its main pruning principles have been extended to remove parameters in the Winograd domain.
\cite{liu2016pruning} removed Winograd-domain kernels, meanwhile, they retained kernels from the original network. However, dimension inconsistency arises since the Winograd-domain kernels are of a higher dimension than the spatial-domain kernels.
%
%To solve this issue, 
\cite{li2017enabling} introduced Winograd layers in exchange for the standard convolutional layers. The pruning and training are simultaneously conducted in the Winograd layers. 
%
%In this fashion, 
Thus, the dimension inconsistency issue is eliminated and the sparsity in Wiongrad domain also increases.
\cite{liu2018efficient} introduced the ReLU operation to the Winograd domain to derive sparse transformed activations. 
%At the same time, 
It improves the possibility of sparse element-wise products in the Winograd domain. 
\cite{yu2019spatial} specified that different locations of the Winograd layers contribute differently to the output activations. 
Despite the progress, these studies lead to hardware-unfriendly irregular sparse patterns, causing imbalanced workloads among the data flows.
To leverage the multiplication reduction from sparsity, \cite{lu2018spwa,yang2020winograd} devised sparse patterns that benefit more from speedups on specialized hardware.

\section{Methodology}

\subsection{3D Winograd}
Giving a 3D convolution kernel $\mathcal{G} \in \mathbb{R}^{C_{o}
\times C_{i} \times r_d \times r_h \times r_w}$ where $C_{o}$ and $C_{i}$ denote the number of output and input channel, and $(r_d, r_h, r_w)$ forms the kernel size,
it is convoluted with a four-dimensional input data $\mathcal{I} \in \mathbb{R}^{C_{i} \times D_{i} \times H_{i} \times W_{i}}$ where $D_{i}$, $H_{i}$ and $W_{i}$ respectively denote the depth, height, and width of the input $\mathcal{I}$.
Typical 3D convolution operation with unit stride and unit dilation can be formulated as:
\begin{equation}\label{conv_vanilla}
    \mathcal{O}{(n,:,:,:)} = \sum_{c=0}^{C_{i}-1} \mathcal{G}{(n,c,:,:,:)}\ast\mathcal{I}{(c,:,:,:)},
\end{equation}
where $\mathcal{O} \in \mathbb{R}^{C_{o} \times D_{o} \times H_{o} \times W_{o}}$ is the output feature map, $D_{o}$, $H_{o}$ and $W_{o}$ denote the depth, height and width of $\mathcal{O}$, respectively, and $\ast$ stands for 3D convolution.

The above 3D operation can be split into multiple basic convolutions, each of which can be optimized to
obtain lower arithmetic complexity by using the Winograd
convolution~\cite{wang2017winograd}. Specifically, the Winograd convolution disassembles the input data $\mathcal{I}$ into several overlapping tiles $\{\mathcal{I}_1, \mathcal{I}_2, ...\}$, in which each tile is a sub-matrix of $\mathcal{I}_k \in \mathcal{I}$ and has a shape of $C_i \times t_d \times t_h \times t_w$. Similar to Eq.\,(\ref{conv_vanilla}), each tile $\mathcal{I}_t$ can be convoluted separately with $\mathcal{G}$, resulting in a basic output tile $\mathcal{O}_t$ that is a sub-matrix of $\mathcal{O}$ and has a shape of $C_{o} \times m_d \times m_h \times m_w$ where $m_d = t_d-r_d+1$, $m_h = t_h-r_h+1$ and $m_w = t_w-r_w+1$.
Note that, any $\mathcal{O}_k$ and $\mathcal{O}_{s}$ are non-overlapping and the spatial convolution result $\mathcal{O}$ can be obtained by reassembling $\{\mathcal{O}_1, \mathcal{O}_2, ...\}$ in order. Eq.\,(\ref{conv_vanilla}) can be further formulated as:

\begin{align}
\label{tiledconv}
 \mathcal{\tilde{I}}
\in \mathbb{R}^{T\times C_{i} \times t_{d} \times t_{h}\times t_{w}} \to \mathcal{\tilde{O}}\in \mathbb{R}^{T\times C_{o} \times m_{d} \times m_{h} \times m_{w}} \nonumber \\
 {\rm via} \quad \mathcal{\tilde{O}}{(k,n,:,:,:)} = \sum_{c=1}^{C_{i}} \mathcal{G}{(n,c,:,:,:)}\ast\mathcal{\tilde{I}}{(k,c,:,:,:)},    
\end{align}
where $\mathcal{\tilde{O}}$, $\mathcal{\tilde{I}}$ denote the disassembled output feature and input data, $\mathcal{\tilde{O}}(k,:,:,:,:)$, $\mathcal{\tilde{I}}(k,:,:,:,:)$ denote basic output tile $\mathcal{O}_k$, basic input tile $\mathcal{I}_k$, respectively, and $T = D_i H_i W_i/(t_d t_h t_w)$ is the number of disassembled tiles. A thorough comprehension can be referred to~\cite{lavin2016fast}.
Notice in what follows, we introduce $r=r_d=r_h=r_w$ and $t = t_d = t_h = t_w$ for brevity since current networks often have a uniform kernel size such as $3\times3\times3$ across different dimensions, and also the basic output tile is characterized with $m = m_{d} = m_{h} = m_{w}$.
% We use $\mathcal{\tilde{O}}\in \mathbb{R}^{T\times C_{o} \times m_{1} \times m_{2} \times m_{3}}$, $\mathcal{\tilde{I}}
% \in \mathbb{R}^{T\times C_{o} \times t_{1} \times t_{2} \times t_{3}}$ to denote 
Since the spatial convolution is computationally intensive, Eq.\,(\ref{tiledconv}) can be further optimized with the Winograd transformation~\cite{lavin2016fast}:
\begin{equation} 
\small
    \mathcal{\tilde{O}}{(k, n, :, :, :)} =  \mathscr{T}_O\Big(\sum_{c=0}^{C_{i}-1}{}\mathscr{T}_K\big(\mathcal{G}(n,c,:,:,:)\big)\odot\mathscr{T}_I(\mathcal{\tilde{I}}{(k,c,:,:,:)}\Big),
\label{WINO3dconv}
\end{equation} 
where $\odot$ stands for the element-wise product.

In Eq.\,(\ref{WINO3dconv}), the kernel $\mathcal{G}{(n, c, :, :, :)}$ and input tiles $\mathcal{\tilde{I}}{(k,c,:,:,:)}$ are individually converted into the Winograd domain of the same shape by the Winograd kernel transformation $\mathscr{T}_K(x) = (KxK^T)^RK^T$ and input transformation $\mathscr{T}_I(x)=(B^TxB)^RB$. Finally, the Winograd-domain kernel and input tile are multiplied in an element-wise manner, the results of which are transformed back to the vanilla spatial domain by the Winograd inverse transformation $\mathscr{T}_O(x)=\big((A^TxA)^RA\big)^R$. 
Herein, $K$, $B$ and $A$ are three transformation matrices determined by $F(m \times m \times m, r \times r \times r)$. Their specific formats can turn to~\cite{lavin2016fast}.
$(\cdot)^R$ denotes clock-wise dimension rotation. Considering a 3-D Tensor $T{(z, y, x)}$, a toy example for $(\cdot)^R$ is illustrated as: $T(z,y,x)^R = T(y,x,z)$.

For better illustration, we first rearrange the spatial kernel $\mathcal{G}\in\mathbb{R}^{C_{o}{\times}C_{i}{\times}r{\times}r{\times}r}$, the input tiles $\mathcal{\tilde{I}}\in\mathbb{R}^{T\times C_{i}{\times}t{\times}t{\times}t}$, and the output tiles $\tilde{\mathcal{O}}\in\mathbb{R}^{T\times C_{o}{\times}m{\times}m{\times}m}$ into 2D matrices $G\in\mathbb{R}^{C_{o}C_{i}{\times}r^3}$,  $\tilde{I}\in\mathbb{R}^{TC_{i}{\times}t^3}$, and $\tilde{O}\in\mathbb{R}^{TC_{o}{\times}m^3}$ accordingly. After rearrangement, Eq.\,(\ref{WINO3dconv}) is modified as a new form:
\begin{equation}
\small
\label{reaWINO3dconv}
  \tilde{O}(kn,:) = \Big(\sum_{c=0}^{C_{i}-1}G(C_{i}n+c, :)T_K\odot {\tilde{I}(kC_i+c,:)T_I}\big)\Big)T_O,
\end{equation}
where $T_K\in\mathbb{R}^{r^3{\times}t^3}$, $T_I\in\mathbb{R}^{t^3{\times}t^3}$,
$T_O\in\mathbb{R}^{t^3{\times}m^3}$ are transformation matrices and the operations with transformation matrices $T_K$, $T_I$, $T_O$ correspond to the process of Winograd input transformation $\mathscr{T}_i(\cdot)$, Winograd kernel transformation $\mathscr{T}_k(\cdot)$ and  Winograd output transformation $\mathscr{T}_o(\cdot)$ in Eq.\,(\ref{WINO3dconv}), respectively\footnote{A comprehensive derivation of Eq.\,(\ref{reaWINO3dconv}) as well as the formats of transformation matrices $T_K$, $T_I$ and $T_O$ can be referred to the appendix.}.
To simplify the description, we denote the Winograd input transformed result $\Tilde{I}T_I$ as $V$ and use $\tilde{\odot}$ to represent the consecutive operations of element-wise product and summation over the output channel in Eq.\,(\ref{reaWINO3dconv}) for ease of the following representation. 
%
%Therefore, 
Eq.\,(\ref{reaWINO3dconv}) can be simplified as:
\begin{equation} \label{aranoutput_fomat}
   \tilde{O} = \big({G}T_K\tilde{\odot}V\big)T_O.
\end{equation}

Particularly, \cite{li2017enabling} introduced a 2D Winograd layer parameterized by a weight tensor $\mathcal{G}_W \in \mathbb{R}^{C_{o} \times C_{i} \times t \times t}$ to replace the Winograd kernel transformation. The element-wise operation costs are expected to decrease from deriving a sparse $\mathcal{G}_W$.
In this paper, we extend the Winograd layer to 3D and introduce the Winograd-domain weight $\mathcal{G}_W \in \mathbb{R}^{C_{o} \times C_{i} \times t \times t \times t}$ to directly perform element-wise products with the Winograd-domain input tiles. We also rearrange $\mathcal{G}_W$ to a 2D Winograd-domain weight matrix $G_W \in \mathbb{R}^{C_{o} C_{i} \times t^3}$, therefore Eq.\,(\ref{aranoutput_fomat}) becomes as follows:
\begin{equation} \label{newoutput_fomat}
   \tilde{O} = \big({G}_W\tilde{\odot}V\big)T_O.
\end{equation} 
The weight matrix $G_W$ is directly inherited from the pre-trained spatial-domain weight matrix transformed into the Winograd domain, \emph{i.e.}, $G_W = G T_K$. After getting $G_W$,  the problem of pruning the 3D Winograd layer is converted into sparsifying the weight matrix $G_W$.

Different from the 1D or 2D Winograd layer, 3D Winograd layer introduces more parameters, which raises a formidable challenge in not only the increasing parameters of sparse training but also accelerating the element-wise product. These two issues are respectively solved in this paper by introducing a low-rank Winograd transformation in Sec.\,\ref{low-rank} and a low-rank oriented sparse granularity in Sec.\,\ref{position-based}, which are also two core contributions of this paper.
In what follows, we refer to the model with Winograd layers as Winograd model and the one with convolutional layers as spatial model.

\subsection{Low-rank Winograd Transformation}\label{low-rank}
%
%Training the Winograd model has a similar pipeline to the spatial model: initializing weights from pretrained model to save overall time, fine-tuning to migrate to downstream task. 
%
%It's not troublesome to obtain the pretrained weights of the Winograd model, since the Winograd-transformed spatial weights can be directly served as the pretrained weights for the Winograd model. Transforming spatial kernels into Winograd kernels drastically increase trainable parameters, yet fine-tuning the full Winograd parameters seems helpless to improve the model's performance, which leads us to the following thoughts: Does the Winograd transformation introduce a part of the noise space? Is that necessary to search in the entire transformed Winograd space for a solution suitable for the downstream task?

%

\begin{figure}
\centering
\subfigure[The proportion of the sum of the top-$i$ singular values to the sum of all.]{
\label{topisin}
\includegraphics[width=0.5\textwidth]{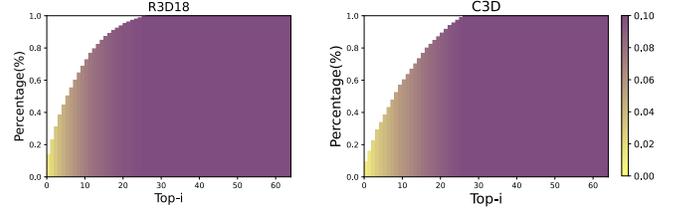} 
}
\subfigure[The proportion of $i$-{th} singular value to the sum of all.]{
\label{ithsin}
\includegraphics[width=0.5\textwidth]{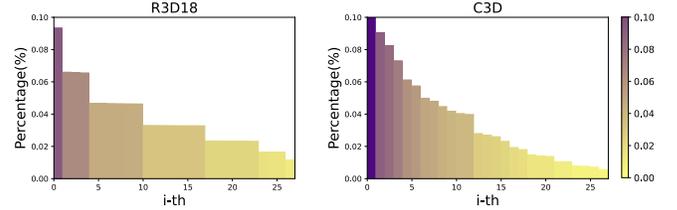} 
}
\caption{Singular value analysis for layer5b of C3D and layer4b$\_$2b of R3D-18.}
\label{vissin}
\end{figure}

The sparse training of the Winograd model shares a two-step pipeline similar to the spatial model, including pruning weights that are unimportant and retraining the model for recovering accuracy.
However, it is challenging to train the 3D Winograd model due to the increasing trainable parameters and expensive element-wise products.
When looking back on the Winograd weight $G_W$, we observe that the over-increasing parameters do not always benefit the performance gains. Then, we analyze the over-increasing parameters by performing singular value decomposition (SVD) on the rearranged Winograd-domain weight matrix $G_W\in\mathbb{R}^{C_oC_i{\times}t^3}$. 
In this fashion, $G_W$ can be represented by the $t^3$-dimensional subspace as: $G_W=\sum_{i=0}^{t^3-1}\sigma_i\vec{u_i}\vec{v_i}^T$, where $\sigma_i$ and $\vec{u_i}$/$\vec{v_i}$ indicate the $i$-th largest singular value and the corresponding left/right singular vector.
Fig.\,\ref{vissin} visualizes the singular values where two phenomena can be observed.
First, among all the $t^3$ singular values, larger-magnitude ones are concentrated in the top-$r^3$ ($r^3=27$ in Fig.\,\ref{vissin}), which is exactly the number of weight elements in the spatial domain. Second, among the top-$r^3$ singular values, those in the front part are much larger than those in the back part.
Such phenomena suggest the existence of over-parameterized weights in the Winograd domain, which means that it may not be necessary to train the Winograd model within the full Winograd domain. A more efficient training way is urgent for 3D Winograd models.
Given the pre-trained Winograd-domain weight $G_W$, we denote the fine-tuned weight as $G_W + \Delta G_W$, where $\triangle{G_W}\in\mathbb{R}^{C_oC_i{\times}t^3}$ denotes the updates from the $G_W$ to the eventual fine-tuned $\bar{G}_W$. 
\cite{hu2021lora} and~\cite{li2022parameter} showed that the update $\Delta G_W$ is supposed to have a low ``intrinsic rank'' if the pre-trained weight $G_W$ is over-parameterized.
This indicates that fine-tuning in the Winograd domain raises attention to the main directions of the whole Winograd space. 
In light of this, we freeze the pre-trained dense Winograd weight $G_W$ first and then achieve low-rank update $\Delta G_W$ by a low-rank decomposition $\Delta G_W = G_{r}G_{c}$ where $G_{r}\in\mathbb{R}^{C_oC_i{\times}s}$ and $G_{c}\in\mathbb{R}^{s{\times}t^3}$ ($s \ll t^3$). 
Therefore, our low-rank Winograd transformation can be finally described as:
% %
% \begin{equation} \label{output_fomat}
%   \mathcal{O}_i = \big(\sum_{c=0}^{C_o-1}{(\mathcal{G}_W +  \mathcal{G}_{r}\mathcal{G}_{c})}_{[c\cdot{C_i}:(c+1){C_i}, :]}\odot{(\mathcal{I}_{i}T_I)}\big)T_O,
% \end{equation} 
% %
%
\begin{equation} \label{lroutput_fomat}
   \tilde{O} = \big((G_W +  G_{r}G_{c})\tilde{\odot}V\big)T_O.
\end{equation} 

%
%Finally, Eq.\,(\ref{fine-tuing-forward}) can be reformulated as:
%\begin{equation}\label{output_2}
%    \bar{\mathcal{O}}_i = \mathscr{T}_O\big((\mathcal{G}_W+\mathcal{G}_{r}\mathcal{G}_{c})\odot\mathscr{T}_I(\mathcal{I}_i)\big).
%\end{equation} 
%
During sparse training, we freeze $G_W$, and upgrade $G_{r}$ and $G_{c}$ only. In this fashion, the amount of trainable parameters are reduced from $C_oC_it^3$ to $C_oC_is + st^3$ in the Winograd domain.
In order to train $G_{r}$ and $G_{c}$ more effectively, we initialize $G_{r}$ by: $G_{r}(:,i)=\alpha\sigma_i\vec{u_i}$ and $G_{c}$ by: $G_{c}(i,:)=\vec{v_i}^T$, where $\alpha$ is a scalar hyper-parameter that controls the amplitude of the update and is set to 0.1 in this paper.
\subsection{Low-Rank Oriented Sparse Granularity}\label{position-based}
\begin{figure*}[t]  %子图加并列
\centering
\includegraphics[width=0.95\textwidth]{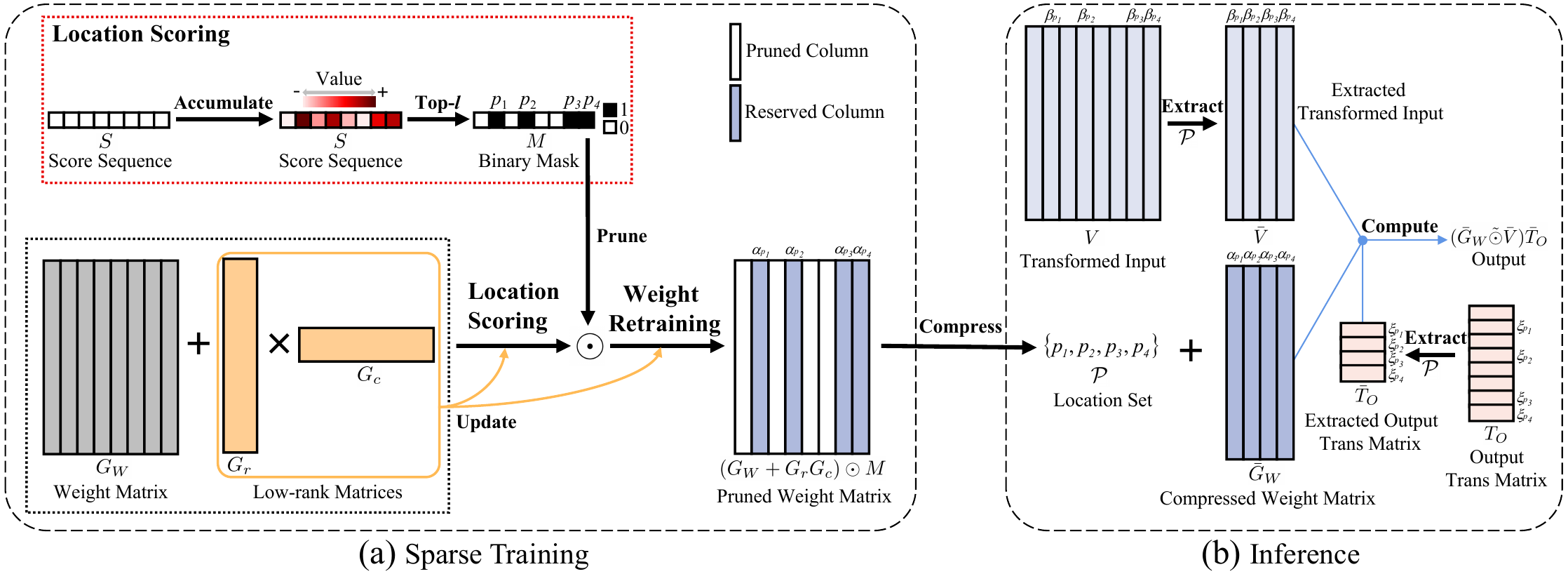} 
\caption{(a)Sparse training of our low-rank oriented sparse granularity. (b)Inference after applying our low-rank oriented sparse granularity.}
\label{pbp}
\end{figure*}
%这一段需要花些篇幅说明winograd kernel更加的position sensitive。
% The Winograd transformation can be modified 
% The pretrained Winograd weights $\mathcal{G}_W$ are transformed from the pretrained weights $\mathcal{G}$ in the spatial domain via $\mathscr{T}_K(\cdot)$ and element-wise products
% and the results for element-wise products in the Winograd doamin needs to be mapped back to the spatial domain through $\mathscr{T}_O(\cdot)$. 
% $XXX$ is also applicable to retrain the sparse model. 
In addition to trainable parameter reduction, we further attempt to decrease the computation cost from the element-wise product at inference time.
By virtue of our low-rank Winograd transformation in Eq.\,(\ref{lroutput_fomat}), the element-wise multiplications can be lessened if most elements in the resulting $G_{W} + G_{r}G_{c}$ are zeros.
\cite{li2017enabling} imposed a sparse constraint upon $G_W$ given that only $G_W$ involves in the element-wise product of Eq.\,(\ref{newoutput_fomat}). 
%
% \cite{yu2019spatial} demonstrated that each output element can be represented by the weighted sum of the products of Winograd weights and transformed input, and further derived an factor matrix, whose coefficient are only determined by $m$ and $t$. The coefficient of factor matrix $\mathbb{R}^{t{\times}t}$ reflects the effect of the weights at different positions on the output. Also, the factor matrix is applied to constrain pruning, which results in very different sparsity at different locations. Compared with the 2D case, the element number of 3D Winograd kernel increase $\frac{t}{r}\times$. Therefore, we assume that
%
However, sparse constraints often cause irregular weight matrix that receives little acceleration, a simple extension of which prevents the practicality in our settings of Eq.\,(\ref{lroutput_fomat}). 

\textbf{Sparse Granularity}.
Instead, in this paper, we devise a low-rank oriented sparsity to purchase effectual speedups. 
Our motive mainly stems from~\cite{yu2019spatial} that measured the element importance of 2D Winograd kernel using a score matrix and removed low-scored weights accordingly which leads to a distinct sparsity at different locations. 
We also intend to measure the weight importance, but at a more regular pattern.
For ease of presentation, we denote our dense 3D Winograd weight as $G_W + G_r G_c = [\alpha_1, \alpha_2, ..., \alpha_{t^3}] \in \mathbb{R}^{C_oC_i \times t^3}$ where each column $\alpha_i \in \mathbb{R}^{C_oC_i \times 1}$. 
Our sparse granularity consists of a single column location in $G_W + G_r G_c$.
In other words, pruning based on column locations results in removing the entire column elements. 

The implementation of our low-rank oriented sparsity is of two stages including location scoring and weight retraining, respectively to filter out pruned target column locations and to recover the pruned model performance.
In the former stage, we freeze the Winograd parameter $G_W$ and initialize trainable parameters $G_r$ and $G_c$ as introduced in Sec.\,\ref{low-rank}. Then, we introduce a score sequence $S \in \mathbb{R}^{t^3}$, which is initialized with zeros, to evaluate location importance. 
Alike to Taylor pruning~\cite{molchanov2019importance}, 
we opt to accumulate the location magnitude and gradient in each training iteration to be served as the values of $S$:
\begin{align}
    S^t = &S^{t-1}+ \frac{1}{C_i^2C_o^2}\big(\sum_{u=0}^{C_iC_o-1}{\left|G_{W}+
 G_{r}G_{c} \right|^{t-1}{(u,:)}}\big)  \nonumber \\ &\odot \big(\sum_{v=0}^{C_iC_o-1}{| \frac{\partial{\mathcal{L}}}{\partial{G_{r}}}\frac{\partial{\mathcal{L}}}{\partial{G_{c}}} |^{t-1}{(v,:)}}\big),
\label{MFML}
\end{align}
where the superscript $t$ represents weight/magnitude and score sequence at the $t$-th training iteration. Here, no additional parameters are introduced during determining $S$.
With the score sequence $S$, we can finally derive a location set $\mathcal{P} = \{p_1, p_2, ..., p_l\}$ that contains locations whose scores within the top-$l$ largest, leading to a sparsity rate of $(t^3-l)/t^3$. Then, we can obtain a binary mask $M \in \mathbb{R}^{t^3}$ as:
%
%Based on scoring sequence, we can finally filter out the location sets $\mathcal{P}$ that contains locations scored in top-$l$, where $\frac{t^3-l}{t^3}$ is the target pruning ratio, and obtain the binary mask $\mathcal{M}$ by: 
% \textbf{Layer-wise}
\begin{equation}
M(i) =
\begin{cases}
 1,  & i\in \mathcal{P},\\
 0,   & \text{Otherwise}.\\
\end{cases}
\label{MFP}
\end{equation} 

The location scoring stage feeds back a fixed binary mask $M$. In the stage of retraining, $M$ is applied to remove low-scored column locations and we only need to fine-tune the trainable parameters $G_r$ and $G_c$ to recover the accuracy of the pruned model.
%
%In the process of Retraining, $M$ is applied to removed selected unimportant columns and we only need to fine-tune the trainble parameters $\mathcal{G}_{r}$ and  $\mathcal{G}_{c}$ to recover the accuracy of the pruned model. 
Therefore, the computation of input tiles becomes:
\begin{equation}\label{computation_1}
 \tilde{O} = \Big(\big((G_W+G_{r}G_{c})\odot{M}\big)\tilde{\odot}{V}\Big)T_O.
\end{equation}

%After retraining, the retrained parameters can be finalized as:
%\begin{equation}
%\mathcal{G}_{W} = (\mathcal{G}_{W}+{\mathcal{G}}_{r}{\mathcal{G}}_{c})\odot{\mathcal{M}}.
%\end{equation}

After retraining, $G_W$ is finally updated by $G_W=(G_W+G_{r}G_{c})\odot{M}=
[\vec{0},\alpha_{p_1},\vec{0},\alpha_{p_2},\cdots,\alpha_{p_l},\vec{0}]$, where $\vec{0}$, $\alpha_{p}$ represent the pruned column and reserved important column, respectively. Fig.\,\ref{pbp}(a) gives an illustrative example of sparse training of our low-rank oriented sparse granularity.

% For model pruned in the filter range, Winograd kernels under each filter are pruned with the same positions and for model pruned in the layer range, Winograd kernels under each layer are pruned with the same positions. Their pruning granularity is $C_o\times{t^3}$ and $t^3$, respectively. Therefore, the former enables higher sparsity, while the latter is more friendly for deployment acceleration. 

%
\textbf{Speedup and Compression Mechanism}.
%
%定义了一个符号$\tilde{\odot}$来表示公式8这种不完全是点积的运算.
%Unlike irregular sparse patterns~\cite{li2017enabling}, the 3D Winograd layer pruned by our proposed $xxx$ result in a very regular sparse pattern, which not only can be compressed simply, but also effectively leverage the multiplication reduction from sparsity. 
%
Unlike the irregular sparse patterns~\cite{li2017enabling}, our low-rank oriented sparse granularity results in a very regular sparse pattern. Therefore, it can well support practical speedups in the inference by simply involving non-zero columns with the multiplication in the code implementation. The model pruned in low-rank oriented sparse granularity can be easily compressed by storing the corresponding reserved columns. The inference process after applying our low-rank oriented sparse granularity is illustrated in Fig.\,\ref{pbp}(b), and its detailed process is provided in the following.

% Unlike the irregular sparse patterns~\cite{li2017enabling}, our low-rank oriented sparse granularity results in a regular, sparse pattern, which is more amenable to practical speedups in inference by allowing the implementation to only consider non-zero columns during multiplication. In addition, the model can be easily compressed by storing only the columns that are retained under this sparsification scheme. We detail it below.

%Same as rearranged $\mathcal{G}_W$, the tranformed input $\mathcal{V}_i\in\mathbb{R}^{C_iC_o\times t^3}$ can also be represented by 
%$\mathcal{V}_i=\mathcal{I}_i\cdot{T_I}=[\beta_1,\beta_2\cdots\beta_{t^3}]$, where $\beta_i\in\mathbb{R}^{C_i\times1}$ denote $i$-th column of $\mathcal{G}_W$, $\mathcal{V}_i$. And output transformation matrix $T_O\in\mathbb{R}^{t^3\times{m^3}}$ can be represented by $T_O=[{\xi_1}^T,{\xi_2}^T\cdots{{\xi_{t^3}}^T}]$, where ${\xi_i}^T$ is the $i$-th row of $T_O$. After pruning, the inference process for our pruned 3D Winograd layer becomes:
Similar to $G_W$, the transformed input tiles can be represented as $t^3$ columns: $V = [\beta_1, \beta_2, ..., \beta_{t^3}]$, where $\beta_i \in \mathbb{R}^{TC_i\times 1}$, and output transformation matrix can be represented as $t^3$ rows: $T_O=[\xi_1; \xi_2;...;\xi_{t^3}]$ where $\xi_i \in \mathbb{R}^{1\times m^3}$. Based on Eq.\,(\ref{newoutput_fomat}), the output can be given as:
\begin{equation}
\begin{aligned}
 \tilde{O}&=
\big([\vec{0},\alpha_{p_1},\vec{0},\alpha_{p_2},\cdots,\alpha_{p_l},\vec{0}]\tilde{\odot}V\big)T_O\\
% &=[\vec{0},\alpha_{p_1}\tilde{\odot}\beta_{p_1},\vec{0},\alpha_{p_2}\tilde{\odot}\beta_{p_l},\cdots,\alpha_{p_l}\tilde{\odot}\beta_{p_l},\vec{0}]T_O\\
&=(\alpha_{p_1}\tilde{\odot}\beta_{p_1})\xi_{p_1}+(\alpha_{p_2}\tilde{\odot}\beta_{p_2})\xi_{p_2}+\cdots+(\alpha_{p_l}\tilde{\odot}\beta_{p_l})\xi_{p_l} \\
& = (\bar{G}_W \tilde{\odot} \bar{V})\bar{T}_O.
\end{aligned}
\end{equation}

Recall that $\mathcal{P} = \{p_1, p_2, ..., p_l\} $ contains column locations with their scores within the top-$l$ largest. Therefore, in the inference stage, we only need to store a compact Winograd weight $\bar{G}_W = [\alpha_{p_1}, \alpha_{p_2}, ..., \alpha_{p_{l}}] \in \mathbb{R}^{C_oC_i \times l}$ as well as the location set $\mathcal{P}$ to extract the columns $\bar{V} = [\beta_{p_1}, \beta_{p_2}, ..., \beta_{p_l}] \in V$ and rows in $\bar{T}_O=[\xi_{p_1}; \xi_{p_2};...;\xi_{p_l}] \in T_O$ which are operated with the corresponding columns in $G_W$. The cost of data extraction is negligible compared to the large percentage of reduction on element-wise products, \emph{i.e.}, $(l/t^3)$. 

\section{Experiments}

\subsection{Experimental Setup}
Our methodology is applied on 3D CNN models including 3D Resnet~\cite{hara2018can} (denoted as R3D) and C3D~\cite{tran2015learning} which consist of plentiful 3D convolution layers with $3\times3\times3$ kernels and a stride of 1. We use pre-trained R3D and C3D on the Kinetics dataset~\cite{carreira2017quo} and Sports-1M dataset~\cite{karpathy2014large}, and fine-tune upon the UCF101 ~\cite{soomro2012ucf101} and HMDB51~\cite{kuehne2011hmdb} datasets, results of which serve as the dense models. 
% The experiments are performed upon the UCF101 ~\cite{soomro2012ucf101} and HMDB51~\cite{kuehne2011hmdb} datasets using pre-trained R3D and C3D on the Kinetics dataset~\cite{carreira2017quo} and Sports-1M dataset~\cite{karpathy2014large}, respectively.
We follow the steps in~\cite{hara2018can} and~\cite{tran2015learning} to generate training samples with 16-frame length, which are cropped to 112 $\times$ 112. For R3D and C3D, we replace all the 3D convolutional layers with $3\times3\times3$ kernels and a stride of 1 (except the first layer) with 3D Winograd layers ($t=4$) and prune on 3D Winograd layers with our proposed low-rank Winograd transformation. The sparsity of 3D Winograd layers is defined as $(t^3-l)/t^3$, where $l$ denotes the number of remaining columns after pruning. We initialize the low-rank matrices $G_r$ and $G_c$ described in Sec.\,\ref{low-rank}. The training epochs are set to 50, of which the first 2 epochs are used for location scoring and the remaining ones are used for weight retraining. The initial learning rate is 1e-4 for C3D and 5e-4 for R3D during location scoring and is divided by 10 for every 15 epochs during weight retraining. Stochastic gradient descent (SGD) serves as the optimizer and cross-entropy loss is adopted to guide model learning.

\subsection{Result Analysis}\label{performance_result}

% \begin{figure*}[t]  %子图加并列
% \centering
% \includegraphics[width=1.00\textwidth]{traincurve.eps} 
% \caption{%Fine-tuning results for dense R3D-based and C3D-based models on UCF101. Full Spatial/Winograd denote fine-tuning with entire sptial/Winograd weights in Spatial/Winograd models. Low rank (L2H) means fine-tuning with Low rank matrices whose ranks are set from low to high.
% Fine-tuning performance comparison on the UCF101 dataset. 
% }
% \label{traincurve}
% \end{figure*}
\textbf{Low-rank Winograd Transformation}. One of the supreme advantages of our low-rank Winograd transformation is that the two less storage-required matrices significantly reduce the trainable parameters during pruning in the Winograd domain. Table\,\ref{worank} compares the results for pruning R3D-18 in our proposed sparse pattern with and without low-rank Winograd transformation on the UCF101 dataset. As can be observed, low-rank Winograd transformation saves numerous trainable parameters ($5.94\times$ in this situation) while achieving better performance when sparsity $<0.7$. In the case of a large proportion (sparsity $>0.7$) of redundant parameters pruned in the Winograd domain, low-rank Winograd transformation can still achieve nearly the same performance as the full Winograd-domain parameters. Results in Table\,\ref{worank} well validate the effectiveness of low-rank Winograd transformation for pruning the redundant Winograd-domain parameters and reducing the storage requirement. 

\begin{table}[!t]
\centering
\caption{Results of our proposed sparse pattern with and without low-rank Winograd transformation. The experiments are conducted using R3D-18 pruned with different sparsity on UCF101.}
\resizebox{1.0\columnwidth}{!}{
\begin{tabular}{c|c|c|c|c|c|c}
\hline
&Trainable  &\multicolumn{5}{|c}{Sparsity}\\
\cline{3-7}
& Parameters &0.1 &0.3 &0.5 &0.7&0.9\\
\hline
w/o Low-Rank& 170.0M &83.3& 83.3 & 82.9& 80.7 & 73.4\\
\hline
Low-Rank& 28.6M &83.4& 83.4& 83.1& 80.7&73.3 \\
\hline
\end{tabular}
}
\label{worank}
\end{table}

\textbf{Low-Rank Oriented Sparse Granularity}. We compare the proposed method with state-of-art methods of pruning 3D CNNs in the different domains, such as FP (Filter Pruning)~\cite{molchanov2016pruning}, RT3D (Group Pruning)~\cite{niu2021rt3d}, DPR (Stripe Pruning)~\cite{zhang2019three} in the spatial domain, and MRP~\cite{qin20223d} in the Winograd domain. Table\,\ref{compwithothers} displays the pruning results of different methods. As can be seen, compared with pruning in 
the Winograd domain, spatial-domain pruning methods suffer the most performance degradation while achieving the same or even lower speedups.  As a contrast, our proposed Winograd-domain pruning pattern achieves the highest speedup ratio and even obtains the highest accuracy for both C3D and R3D-18. For example, when pruning C3D, our method achieves $5.8\times$ speedups with an accuracy loss of $0.9\%$ while MRP achieves $4.3\times$ speedups with an accuracy loss of $1.1\%$, and RT3D achieves only $3.6\times$ speedups with an accuracy loss of $1.4\%$. Similarly, our method also outperforms other methods by a large margin when pruning R3D-18.

\begin{figure}[t]  %子图加并列
\centering
\includegraphics[width=0.48\textwidth,trim={70 1 90 20}]{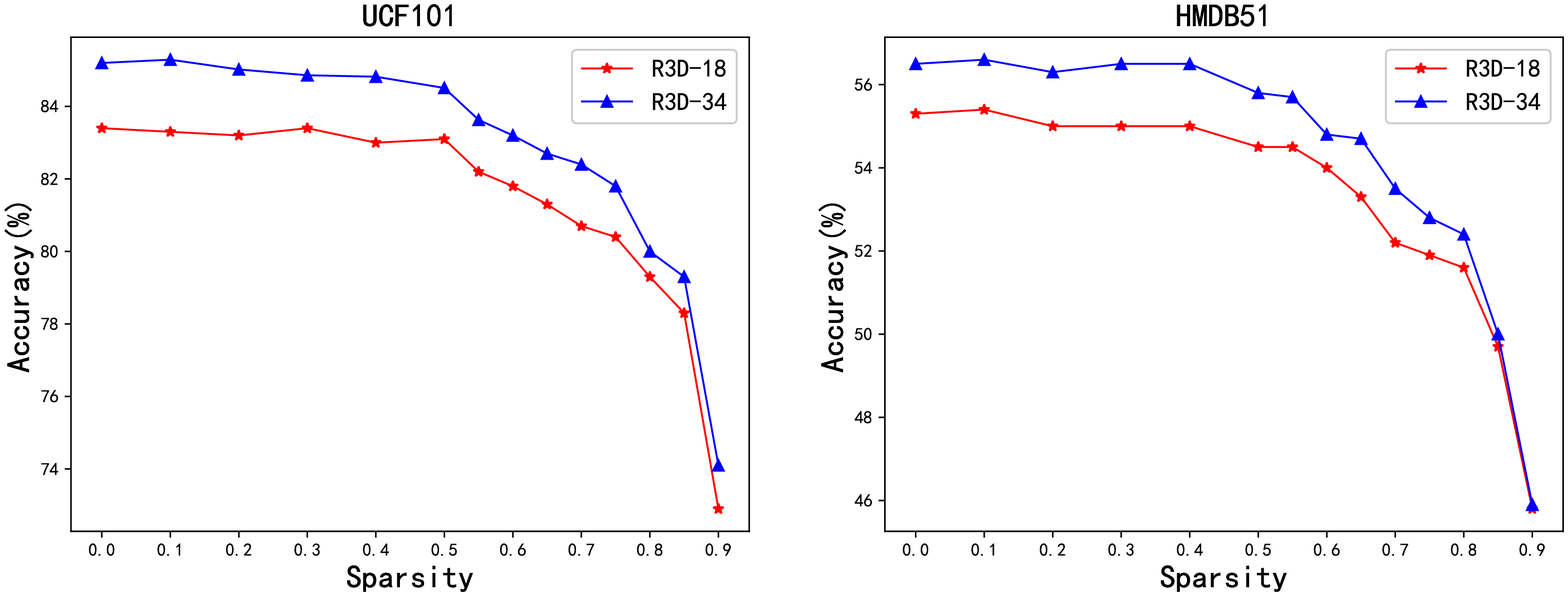}
\caption{Results of R3D-18 and R3D-34 pruned with our proposed low-rank oriented sparse granularity under different sparsity. The experiment is conducted on UCF101(left) and HMDB51(right).}
\label{accspar}
\end{figure}

To further investigate the effect of sparsity on our proposed method, we evaluate low-rank oriented sparse granularity on R3D-18 and R3D-34 on UCF101 and HMDB51 datasets. Fig.\,\ref{accspar} demonstrates the performance results of R3D-18 and R3D-34 under different sparsity ratios. As can be seen, our method is capable of maintaining performance drops within $1\%$ when the sparsity ratio $<0.5$. However, due to the extremely regular sparse granularity of our pruning pattern, our method degrades drastically if the sparsity ratio $>0.8$.

\begin{table}[t]
\centering
\caption{Results of different pruning methods on UCF101, where * denotes our re-implementation. The experiment is conducted using C3D (baseline accuracy 81.6$\%$) and R3D-18  (baseline accuracy 83.5$\%$) and the speedup ratio is computed by GFLOPs reduction.}
\resizebox{1.0\columnwidth}{!}{
\begin{tabular}{c|c|c|c}
\hline
Model & Methods & Speedup & Accuracy(\%)$\downarrow$\\
\hline
\multirow{10}{*}{C3D}&FP*~\cite{molchanov2016pruning}&1.6$\times$&1.6\\
&FP*~\cite{molchanov2016pruning}&3.7$\times$&10.1\\
&DPR~\cite{zhang2019three}&2.0$\times$&3.3\\
&DPR~\cite{zhang2019three}&4.0$\times$&6.6\\
&RT3D~\cite{niu2021rt3d}&2.6$\times$&1.0\\
&RT3D~\cite{niu2021rt3d}&3.6$\times$&1.4\\
&MRP~\cite{qin20223d}&3.8$\times$&0.4\\
&MRP~\cite{qin20223d}&4.3$\times$&1.1\\
&Ours(sparsity=0.3)&4.3$\times$&0.6 \\
&Ours(sparsity=0.5)&5.8$\times$&0.9 \\
\hline
\multirow{7}{*}{R3D-18}&FP*~\cite{molchanov2016pruning}&1.6$\times$&2.6\\
&FP*~\cite{molchanov2016pruning}&4.0$\times$&8.5\\
&DPR*~\cite{zhang2019three}&2.0$\times$&3.4\\
&DPR*~\cite{zhang2019three}&4.0$\times$&5.0\\
&MRP~\cite{qin20223d}&3.2$\times$&0.1\\
&MRP~\cite{qin20223d}&3.8$\times$&1.2\\
&Ours(sparsity=0.3)& 4.3$\times$&0.1\\
&Ours(sparsity=0.5)& 5.0$\times$&0.5\\
\hline
\end{tabular}
}
\label{compwithothers}
\end{table}

\textbf{Acceleration Performance}. The acceleration capacity of our proposed method is evaluated on CPUs-based platforms. We first compare our proposed method with Img2col and Winograd algorithms which are commonly used methods for accelerating convolutional operations. For fair comparison, the inference code for our method and the compared methods are all implemented based on the mobile inference framework of Tencent ncnn\footnote{https://github.com/Tencent/ncnn.} and optimized by advanced SIMD (Single Instruction, Multiple Data). We respectively deploy C3D with different acceleration methods to obtain the end-to-end network inference latency on the Redmi Note10 platform. Table\,\ref{speedtest} presents the experimental results. 
Compared with Img2col, the Winograd algorithm reduces a large amount of multiplications, which makes it obtain a 2.1$\times$ acceleration ratio and our method is able to reduce more inference latency based on the Winograd algorithm. For example, our method obtains a 5.0$\times$ acceleration ratio under the sparsity of 0.9 and a 3.4$\times$ acceleration ratio under the sparsity of 0.3. The results shown in Table\,\ref{speedtest} demonstrate the excellent acceleration capacity of our proposed method.

\begin{figure}[t]  %子图加并列
\centering
\includegraphics[width=0.48\textwidth,trim={70 1 90 20}]{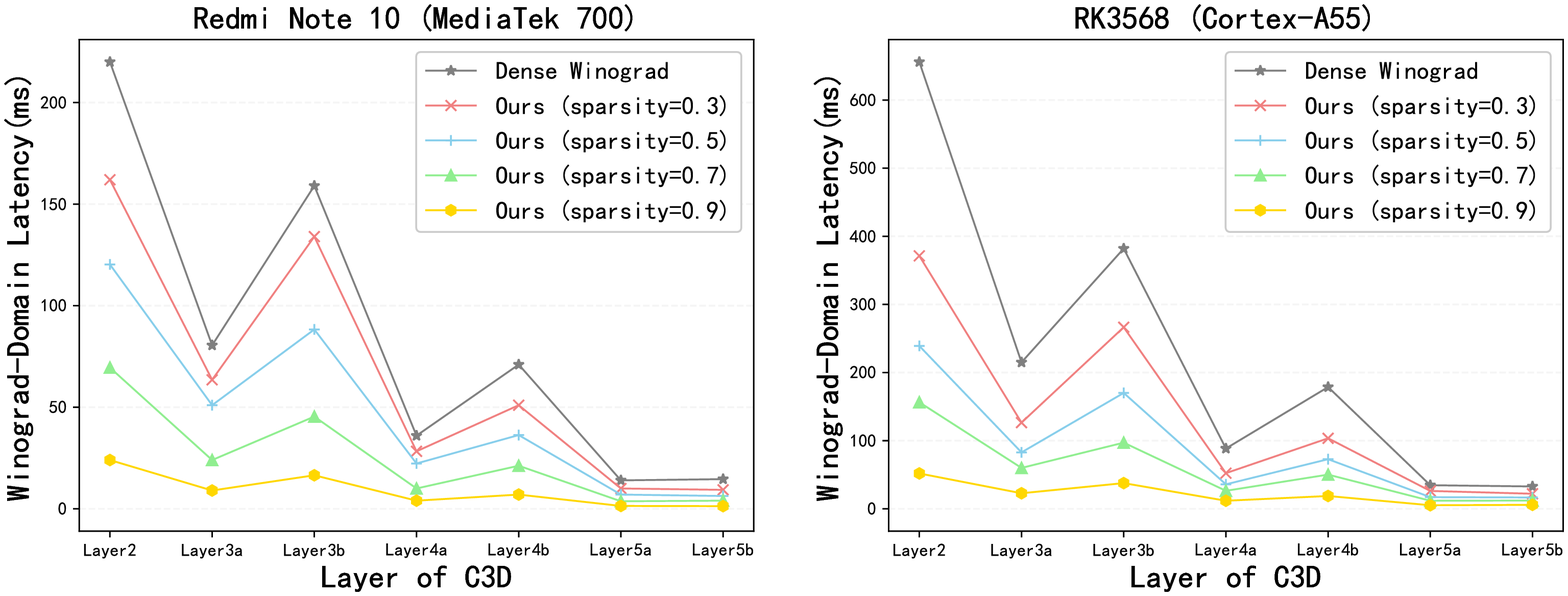}
\caption{Comparison of the inference latency in Winograd domain between our low-rank Winograd convolution and dense Winograd convolution. The experiment is conducted using different layers of C3D on the mobile phone of Redmi Note 10 equipped with a MediaTek 700 CPU (left), and the computing platform of RK3568 equipped with Cortex-A55 CPUs (right).}
\label{acceler}
\end{figure}

For Winograd convolution, the element-wise products in the Winograd domain occupy a vast majority of the inference time. The core acceleration mechanism of our method is to reduce a large number of operations in the Winograd domain by pruning Winograd-domain weights. Due to our proposed regular sparse pattern, extracting the corresponding arithmetic data only introduces a very small amount of overhead which makes sure that the sparsity obtained by pruning can be converted into actual speedups. Fig.\,\ref{acceler} manifests the Winograd-domain inference latency of our method and its dense Winograd counterpart. As can be seen from the table, our proposed pruning pattern effectively translates the sparsity into actual speedups in the Winograd domain across different layers of C3D.

\begin{table}[!t]
\centering
\caption{Inference latency comparison among different acceleration strategies. The experiment is conducted using C3D on the platform of Redmi Note 10 equipped with  MediaTek 700 CPUs.}
\resizebox{1\columnwidth}{!}{
\begin{tabular}{c|c|c|c|c}
\hline
Acceleration &\multirow{2}{*}{Sparsity} &\multirow{2}{*}{Accuracy(\%)}  &\multirow{2}{*}{CPU(ms)} &Acceleration\\
Strategy& &&&Ratio\\
\hline
Img2col&-&81.6&2970&-\\
\hline
Winograd&-&81.6&1211&2.5$\times$\\
\hline
\multirow{4}{*}{Ours}&0.0&81.6&1144&2.6$\times$\\
&0.3&81.0&873&3.4$\times$\\
&0.5&80.7&789&3.8$\times$\\
&0.7&79.8&715&4.2$\times$\\
&0.9&70.6&594&5.0$\times$\\
\hline
\end{tabular}
}
\label{speedtest}
\end{table}

\subsection{Ablation Study}

\textbf{Rank Selection}. To further explore the effect of ranks on low-rank Winograd transformation, we have tried different settings of rank $s$ for $G_r$ and $G_c$. For R3D, we set ranks based on different blocks, and for C3D, we set ranks based on different layers. Specifically, for a rank set $\mathbb{S}=\{s_1,\cdots,s_l\}$, $s_i$ denotes the concrete rank for $i$-th Winograd layer or $i$-th block containing Winograd layer and $l$ denotes the total number of Winograd layers/blocks. Table\,\ref{rank-select} shows the effect of rank setting. As can be observed from the displayed table, a modest increase in ranks will improve the performance of low-rank Winograd transformation and deeper layers tend to require larger ranks than shallow layers. Considering the overall effect, we finally choose
$\mathbb{S}=\{2,4,8,12\}$ for R3D model and $\mathbb{S}=\{1,1,2,4,8,12,12\}$ for C3D model to perform our experiments in this paper.

\begin{table}[t]
\centering
\caption{Comparison of performance on R3D-18 and C3D pruned with different rank settings (with sparsity $=0.75$), where $\star$ denotes our chosen rank setting. The experiment is conducted on UCF101.}
\resizebox{1.0\columnwidth}{!}{

\begin{tabular}{cccc}
    \toprule
   \multirow{2}{*}{Model} & \multirow{2}{*}{Rank Setting}   & Trainable & \multirow{2}{*}{Accuracy (\%)} \\
   & &Parameters&\\
    \midrule 
    \multirow{5}{*}{R3D-18} & \{2,2,2,2\} & 5.3M    &80.0 \\
     &  \{8,8,8,8\}  & 21.3M  &80.4    \\
     &  \{12,12,12,12\}  & 31.9M   &80.4   \\
     & \{12,8,4,2\} &  7.4M  &80.2  \\
     & \{2,4,8,12\}$\star$ &  28.6M  &80.6\\
    \midrule
\multirow{5}{*}{C3D} &  \{2,2,2,2,2,2,2\}&    7.8M  &78.2\\
&\{8,8,8,8,8,8,8\}&    31.3M  &78.4\\
 &  \{12,12,12,12,12,12,12\}& 46.9M&79.3\\
     &  \{12,12,8,4,2,1,1\}&  9.9M &77.9 \\
    &  \{1,1,2,4,8,12,12\}$\star$ & 34.7M &79.3  \\

    \bottomrule
\end{tabular}
}

\label{rank-select}
\end{table}

\textbf{Indicators of Location Importance}. In the stage of location scoring, the score sequence $S$ updated by different metrics will produce different pruned columns, result of which greatly affects the performance of the sparse model. We compare several different indicators on C3D and R3D-18 models in Table\,\ref{scoring-seq}. The results show that when selecting pruned columns, gradient-based score ($S_{grad}$) has a greater impact on the assessment of selecting locations than magnitude-based score ($|G_W|$ and $|G_rG_c|$), while the combination of the two gives the best results.

\begin{table}[t]
\caption{
Comparison of performance on R3D-18 and C3D pruned with different indicators of location importance (with sparsity $=0.75$), where $S_{grad}=\frac{\partial{\mathcal{L}}}{\partial{G_{r}}}\frac{\partial{\mathcal{L}}}{\partial{G_{c}}}$ and $\star$ denotes our chosen indicator. The experiment is conducted on UCF101.}
\centering
\resizebox{0.7\columnwidth}{!}{
\begin{tabular}{cccccc}
    \toprule
     Indicator   &  C3D & R3D-18\\
    \midrule  
     $|G_{r}G_{c}|$  &74.5& 73.6\\
     $|G_W+G_{r}G_{c}|$  & 76.6 & 76.9\\
     $|S_{grad}|$ & 78.5&79.8\\
     $|G_{r}G_{c}|\odot |S_{grad}| $& 78.4 & 80.4\\
     $|G_W+G_{r}G_{c}|\odot|S_{grad}|\star$ & 79.2 & 80.6\\
    \bottomrule 
\end{tabular}
}

\label{scoring-seq}
\end{table}

\section{Conclusion}
Here, we have presented a novel low-rank Winograd transformation to reduce the over-parameterized issue in 3D CNNs. We decouple the original Winograd weight matrix into two less storage-required matrices, leading to remarkable trainable parameter reduction. The low-rank constraint well eliminates the redundant parameters and drives the updating toward the main directions of the whole Winograd space. Consequently, our low-rank Winogrard transformation leads to a better performance increase.
In addition, we have also introduced a low-rank oriented sparse granularity to purchase effectual speedups. It models the column-wise importance of the Winograd weight matrix and removes the low-scored ones. In this fashion, the sparsity tends to be more regular, which therefore better supports the practical acceleration in comparison with the existing irregular sparsity.

%% The file named.bst is a bibliography style file for BibTeX 0.99c
\bibliographystyle{named}
\bibliography{ijcai23}

\appendix
\section{Derivations of Eq.\,\ref{reaWINO3dconv} and Transformation Matrices}
To derive Eq.\,\ref{reaWINO3dconv}, we need to derive the form of the Winograd transformations after rearranging $\mathcal{G}$, $\mathcal{\Tilde{I}}$. We first introduce the conclusion, after rearranging $\mathcal{G}$, $\mathcal{\Tilde{I}}$, the Winograd transformations $\mathscr{T}_K(\cdot)$, $\mathscr{T}_I(\cdot)$, and $\mathscr{T}_O(\cdot)$ are supposed to be modified accordingly:

\begin{align}
    &\mathscr{T}_K(\mathcal{G})=(K\mathcal{G}K^T)^RK^T, \,\mathcal{G}\in\mathbb{R}^{C_o{\times}C_i{\times}r{\times}r{\times}r}\to \nonumber \\
    &\mathscr{T}_K(G)=GT_K, \,G\in\mathbb{R}^{C_oC_i{\times}r^3},   \label{mWk}
    \\&
    \mathscr{T}_I(\mathcal{\Tilde{I}})=(B^T\mathcal{\Tilde{I}}B)^RB, \, \mathcal{\Tilde{I}}\in\mathbb{R}^{T{\times}C_i{\times}t{\times}t{\times}t}\to  \nonumber\\
    &\mathscr{T}_I(\Tilde{I})=\Tilde{I}T_I, \,\Tilde{I}\in\mathbb{R}^{TC_i{\times}t^3}, \label{mWi}  
    \\&
    \mathscr{T}_O(\mathcal{V})=\big((A^T\mathcal{V}A)^RA\big)^R, \, \mathcal{V}\in\mathbb{R}^{T\times C_o{\times}t{\times}t{\times}t}\to \nonumber\\
     &\mathscr{T}_O(V)=V{T_O}, \,V\in\mathbb{R}^{TC_o{\times}t^3},
\label{mWo}
\end{align}
where $T_K$, $T_I$, and $T_O$ are transformation matrices that we need to derive.

We start with the 2D-version. Giving a convolution weight $\textsl{g}\in \mathbb{R}^{{r}\times{r}}$ and it is transformed into Winograd weight $\textsl{g}_W \in \mathbb{R}^{{t}\times{t}}$ by $\textsl{g}_W=K\textsl{g}K^T$. Here, we introduce an equation derived by ~\cite{yu2019spatial}:
\begin{equation}\label{rwwbcw}
    \textsl{g}_W(j,k) = \sum_{v=0}^{r-1}{\sum_{w=0}^{r-1}K{(j,v)}{K{(k,w)}}\textsl{g}{(v,w)}}.
\end{equation} 
Eq.\,(\ref{rwwbcw}) indicates that each element of the Winograd weight can be represented by elements of the convolution weight. This conclusion still holds in the 3D case.

Back to the 3D-version, the 3D convolution weight $\mathcal{G} \in \mathbb{R}^{{r}\times{r}\times{r}}$ is transformed into Winograd weight $\mathcal{G}_W \in \mathbb{R}^{{t}\times{t}\times{t}}$ by $(K\mathcal{G}K^T)^R K^T$.
We further divide $(K\mathcal{G}K^T)^RK^T$ into three steps: $\mathcal{Q}=K\mathcal{G}K^T$,$\hat{\mathcal{Q}}=(\mathcal{Q})^R$, and $\mathcal{G}_W=\hat{\mathcal{Q}}K^T$. 
For $\mathcal{Q}=K\mathcal{G}K^T$, we have:
\begin{equation}\label{deriv1}
    \mathcal{Q}{(i,j,k)} = \sum_{v=0}^{r-1}{\sum_{w=0}^{r-1}K{(j,v)}{K{(k,w)}}\mathcal{G}{(i,v,w)}},
\end{equation}
where $0\leq{j,k}\leq{t-1}$, $0\leq{i}\leq{r-1}$. Then we rotate $\mathcal{Q}$ clockwise to $\hat{\mathcal{Q}}$, element of which can be further represented by:
\begin{equation}\label{deriv2}
    \hat{\mathcal{Q}}{(j,k,i)} = \sum_{v=0}^{r-1}{\sum_{w=0}^{r-1}K{(j,v)}{K{(k,w)}}\mathcal{G}{(i,v,w)}}.
\end{equation}
After that, each element of $\mathcal{G}_W$ can be calculated by elements in $\hat{\mathcal{Q}}$ and $G$:
\begin{equation}\label{deriv4}
\begin{aligned}
 \mathcal{G}_W{(x,y,z)}&= \sum_{u=0}^{r-1}K{(z,u)}\hat{\mathcal{Q}}{(x,y,u)} \\
&=\sum_{u=0}^{r-1}\sum_{v=0}^{r-1}{\sum_{w=0}^{r-1}K{(x,v)}{K{(y,w)}{K{(z,u)}}\mathcal{G}{(u,v,w)}}},
\end{aligned}
\end{equation}
where $0\leq{x,y,z}\leq{t-1}$.
We then rearrange $\mathcal{G}$ and $\mathcal{G}_W$ into vectors:
\begin{equation}\label{deriv5}
\begin{aligned}
&\mathcal{G}\in{\mathbb{R}^{{r}\times{r}\times{r}}}\to{G=[a_1,a_2,\cdots,a_{r^3}],G\in{\mathbb{R}^{1\times{r^3}}}},\\
&\mathcal{G}_{W}\in{\mathbb{R}^{{t}\times{t}\times{t}}}\to{G_W=[b_1,b_2,\cdots,b_{t^3}],G_{W}\in{\mathbb{R}^{1\times{t^3}}}}.
\end{aligned}
\end{equation}
%where each position of the vector corresponds to each element of $\mathcal{G}$ and $\mathcal{G}_W$ before the rearrangement.

Let us describe Eq.\,(\ref{deriv4}) in another way where each position in $\mathcal{G}_W$ ($G_W$) can be calculated by combination of coefficients of positions in $\mathcal{G}$ ($G$):
\begin{equation}\label{deriv6}
b_i=a_1c_{1i}+a_2c_{2i}+\cdots+a_{r^3}c_{{r^3}i}.
\end{equation}
Therefore, $G_W$ and $G$ can be related by matrix multiplication:
\begin{equation}\label{deriv8}
[b_1,b_2,\cdots,b_{t^3}]=[a_1,a_2,\cdots,a_{r^3}]\cdot\begin{bmatrix}
c_{11} & c_{12}&\cdots& c_{1t^3}\\
c_{21} & c_{22}&\cdots& c_{2t^3}\\
\vdots&\vdots &\ddots &\vdots\\
c_{r^31}&c_{r^32} & \cdots&c_{r^3t^3} \\
\end{bmatrix},
\end{equation}
and it can be further abbreviated as:
\begin{equation}\label{deriv7}
G_W=\mathscr{T}_{K}(G)=GT_K, \;
T_K=\begin{bmatrix}
c_{11} & c_{12}&\cdots& c_{1t^3}\\
c_{21} & c_{22}&\cdots& c_{2t^3}\\
\vdots&\vdots &\ddots &\vdots\\
c_{r^31}&c_{r^32} & \cdots&c_{r^3t^3} \\
\end{bmatrix}.
\end{equation}
Combining with Eq.\,(\ref{deriv4}), each element of $T_K$ (kernel transformation matrix) can be calculated as follows:

% \begin{equation}\label{deriv9}
% \sum_{0\leq{x,y,z}\leq{t-1}}{\sum_{0\leq{u,v,w}\leq{r-1}}{T_{K[r^2u+r{v}+w,t^2x+t{y}+z]}}}=\sum_{0\leq{x,y,z}\leq{t-1}}{\sum_{0\leq{u,v,w}\leq{r-1}}{G_{[x,v]}\cdot{G_{[y,w]}}\cdot{G_{[z,u]}}}}
% \end{equation}
\begin{equation}\label{deriv9}
\begin{aligned}
T_K(i,j)&={K{(x,v)}\cdot{K{(y,w)}}\cdot{K{(z,u)}}},\\
&i=r^2u+r{v}+w, \\
&j=t^2x+t{y}+z,\\
\end{aligned}
\end{equation}
where $0\leq{i}\leq{r^3-1}$, $0\leq{j}\leq{t^3-1}$, $0\leq{u,v,w}\leq{r-1}$ and $0\leq{x,y,z}\leq{t-1}$.

So far, we have complemented the derivation of Eq.\,(\ref{mWk}). The above process can be also applied to derive Eq.\,(\ref{mWi}) and Eq.\,(\ref{mWo}) and acquire transformation matrices $T_I$ and $T_O$. Finally, we obtain Eq.\,\ref{reaWINO3dconv} by substituting Eq.\,(\ref{mWk}), Eq.\,(\ref{mWi}), and Eq.\,(\ref{mWo}) into Eq.\,(\label{WINO3dconv}).

\section{Low-Rank Winograd Transformation for Dense Training}

We also conduct experiments to evaluate the performance of low-rank transformation for dense training. The experiment is conducted by pre-training R3D and C3D on the Kinetics dataset~\cite{carreira2017quo} and Sports-1M dataset~\cite{karpathy2014large}, and fine-tuning upon the UCF101 via different training methods. We compare the results of fine-tuning upon the full spatial domain, vanilla full Winograd domain, and our proposed low-rank Winograd domain.

\begin{figure}[h]
\subfigure[R3D-18]{
\label{denser3d}
\includegraphics[scale=0.30,trim={30 1 20 50}]{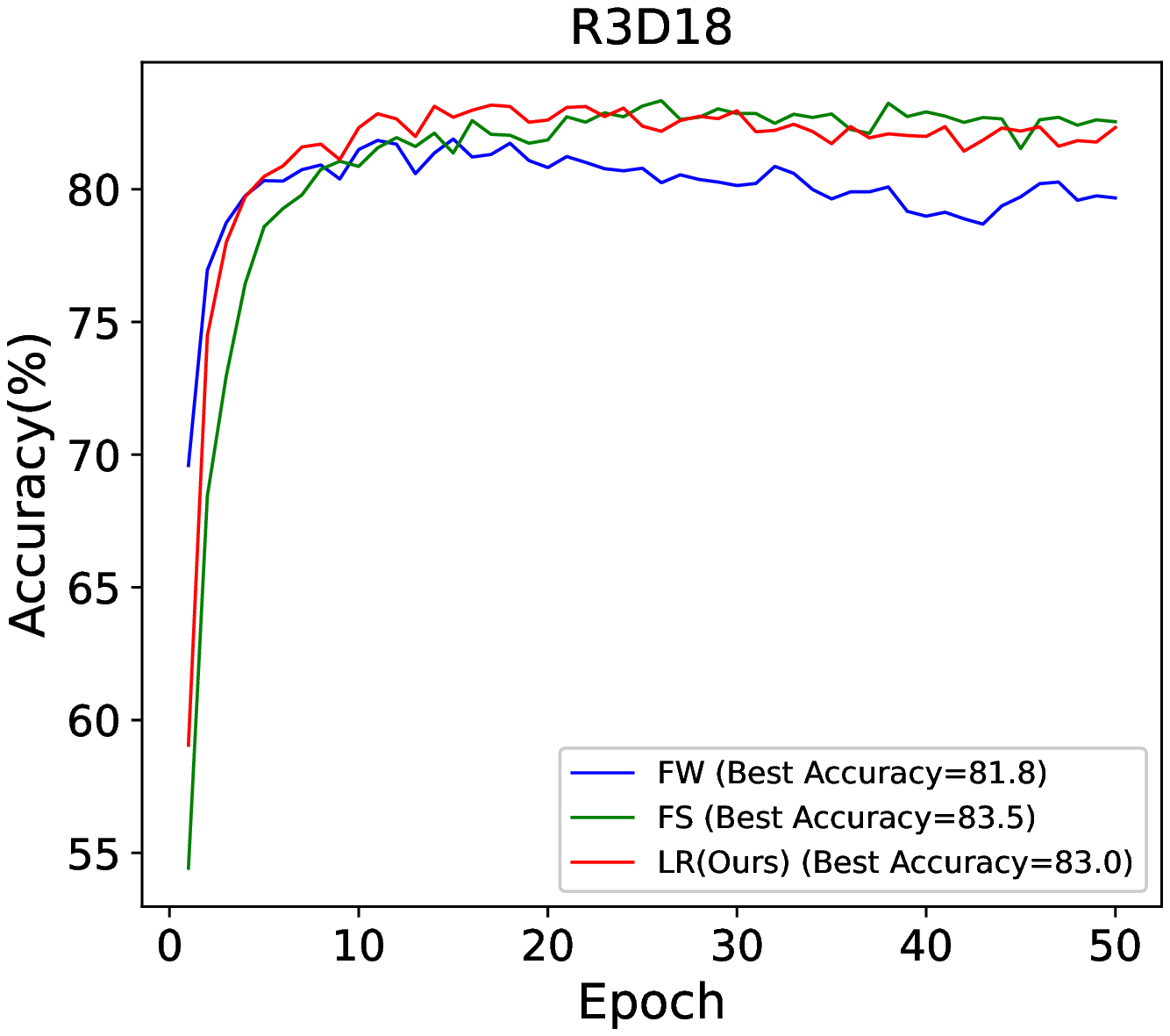} \label{ftr3d}
}
\subfigure[C3D]{
\includegraphics[scale=0.30,trim={20 1 40 50}]{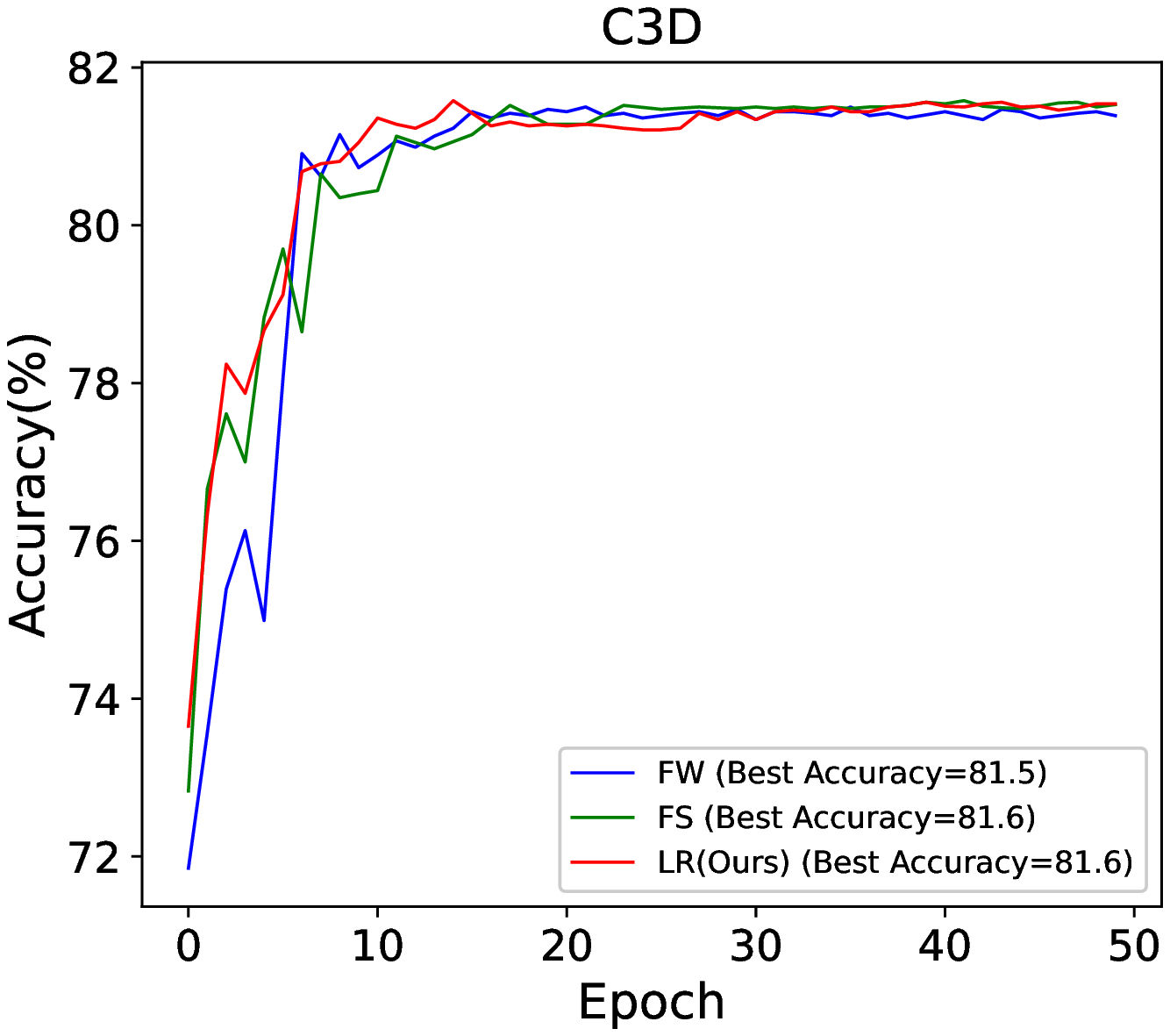}
}
\caption{%Fine-tuning results for dense R3D-based and C3D-based models on UCF101. Full Spatial/Winograd denote fine-tuning with entire sptial/Winograd weights in Spatial/Winograd models. Low rank (L2H) means fine-tuning with Low rank matrices whose ranks are set from low to high.
Comparison among fine-tuning in the full Winograd domain (FW), fine-tuning in the full spatial domain (FS) and fine-tuning in the low-rank Winograd domain (LR) on R3D18 and C3D. The experiment is conducted on the UCF101 dataset.
}
\label{denselr}
\end{figure}

Fig.\,\ref{denselr}(a) shows the fine-tuning results on R3D18. As can be seen, the vanilla Winograd transformation performs the worst almost across the whole fine-tuning stage. Quantitatively, it causes 1.7\% accuracy drops in the end, which well demonstrates our claim that more parameters from Winograd transformation do not always benefit model capacity but cause model redundancy. 
In contrast, our low-rank Winograd transformation manifests supreme performance in comparison with the vanilla version, even on par with the spatial model. The performance gains mostly come from the fact that we drive weight updating towards the main directions of the whole Winograd space.
Fig.\,\ref{denselr}(b) continues the results on C3D. Similar to R3D, the vanilla Winograd transformation suffers the most performance drops, around 0.1\% over the spatial model. On the contrary, by removing the redundancy, our low-rank transformation achieves the same performance as the spatial model. These results again demonstrate the value and the feasibility of our method.

\section{The Validity of Low-Rank Winograd Transformation}
In this section, we analyze the validity of low-rank Winograd transformation. We fine-tune C3D in the vanilla full Winograd domain on UCF101. After fine-tuning, we can get the updates from the pre-trained weight $G_W$ to the eventual fine-tuned $\bar{G}_W$ by $\triangle{G_W}=\bar{G}_W-G_W$. And then we perform singular value decomposition on $\triangle{G_W}$. By this way, $\triangle{G_W}$ can be represented by $t^3$ subspaces:
$\triangle{G_W}=\sum_{i=0}^{t^3-1}\triangle{\sigma_i}\triangle{\vec{u_i}}\triangle{\vec{v_i}}^T$, where $\triangle{\sigma_i}$, $\triangle{\vec{u_i}}$/$\triangle{\vec{v_i}}$ are the $i$-th singular value, left/right singular vector of $\triangle{G_W}$. The magnitude of $\triangle{\sigma_i}$ can be regarded as the importance of the subspace $\triangle{\vec{u_i}}\triangle{\vec{v_i}}^T$. 

\begin{figure}[t]  %子图加并列
\centering
\includegraphics[width=0.30\textwidth,trim={70 1 90 35}]{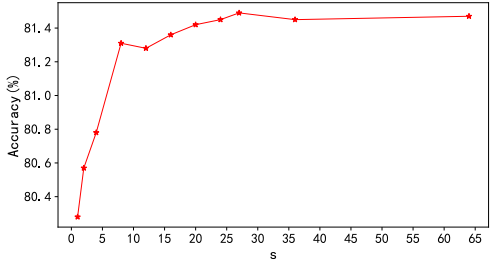}
\caption{The accuracy of C3D on UCF101 when only the top-$s$ subspaces of $\triangle{G_W}$ are retained.}
\label{accremains}
\end{figure}

To explore how the model performance would be affected if only a few parts of the subspace were retained, we directly test the accuracy of the model by adding $\sum_{i=0}^{s-1}\triangle{\sigma_i}\triangle{\vec{u_i}}\triangle{\vec{v_i}}^T$ to the pre-trained weight $G_W$, where $s$ is the number of reserved subspaces. The result is shown in Fig.\,\ref{accremains}. As can be observed, the accuracy of the model does not decrease significantly until $s<8$, when $s$ reduces from 64 to 27, the model accuracy even increases. This suggests that a large part of the space introduced by the Winograd transformation may have hindered the training of the Winograd model. The training process only needs to focus on a small portion of the subspaces, which just fits our proposed low-rank Winograd transformation.

% \begin{table}[htbp]
% \caption{The accuracy of C3D on UCF101 when only the top-$r$ subspaces of $\triangle{\mathcal{G}_W}$ are retained.}
% \centering
% \resizebox{0.9\columnwidth}{!}{
% \begin{tabular}{cccccccccccc}
%     \toprule
%      Rank &  64 &36 &27 &24 & 20& 16& 12& 8& 4 &2 &1 \\ 
%         \midrule 
%      Accuracy (\%) &81.47&81.45&81.49&81.45&81.42 & 81.36&81.28 & 81.31 & 80.78& 80.57& 80.28 \\
%     \bottomrule
% \end{tabular}

% }

% \label{rank-de}
% \end{table}

\end{document}